\documentclass[conf]{new-aiaa}
\usepackage[utf8]{inputenc}

\usepackage{graphicx}
\usepackage{epsfig} 
\usepackage{amsmath}

\usepackage{bm}
\usepackage{booktabs}
\usepackage{subcaption}
\usepackage[version=4]{mhchem}
\usepackage{siunitx}
\usepackage{longtable,tabularx}
\usepackage{times} 
\usepackage{xcolor}
\definecolor{myblue}{HTML}{1383bf}
\definecolor{myred}{HTML}{FF5757}
\definecolor{myblue-1}{HTML}{035362}
\definecolor{myblue0}{HTML}{096F82}
\definecolor{myblue1}{HTML}{0097B2}
\definecolor{myblue2}{HTML}{07B3D1}
\definecolor{myblue3}{HTML}{39C2DB}
\hypersetup{bookmarksopen,bookmarksnumbered,
pdfpagemode=UseOutlines,
colorlinks=true,
linkcolor=myred, 
anchorcolor=blue,
citecolor=myblue, 
filecolor=blue,
menucolor=blue,
urlcolor=myblue-1 
}
\newcommand{\hoa}[1]{\textcolor{black}{#1}} 

\newcommand{\ModelName}{\textit{UniSaT}}
\newcommand{\ModelNameWithSpace}{\textit{UniSaT }}

\title{\ModelName: Unified-Objective Belief Model and Planner to Search for and Track Multiple Objects}

\author{Leonardo Santos\footnote{Co-first author, Undergraduate Student, School of Engineering, {{\tt\footnotesize leohmcs@ufmg.br}}.}}
\affil{Federal University of Minas Gerais, Belo Horizonte, MG, Brazil}
\author{Brady Moon\footnote{Co-first author, Ph.D Student, Robotics Institute, School of Computer Science, {\tt\footnotesize  bradym@andrew.cmu.edu}.} and Sebastian Scherer\footnote{Associate Research Professor, Robotics Institute, School of Computer Science, {\tt\footnotesize  basti@andrew.cmu.edu}.}}
\affil{Carnegie Mellon University, Pittsburgh, PA, USA}
\author{Hoa Van Nguyen\footnote{Lecturer, Department of Electrical and Computer Engineering, {\tt\footnotesize  hoa.v.nguyen@curtin.edu.au}.}}
\affil{Curtin University, Bentley, WA 6102, Australia}

\begin{document}

\maketitle

\begin{abstract}
Path planning for autonomous search and tracking of multiple objects is a critical problem in applications such as reconnaissance, surveillance, and data gathering. 
Due to the inherent competing objectives of searching for new objects while maintaining tracks for found objects, most current approaches rely on multi-objective planning methods, leaving it up to the user to tune parameters to balance between the two objectives, usually based on heuristics or trial and error.
In this paper, we introduce \ModelNameWithSpace (\textit{Unified Search and Track}), a novel unified-objective formulation for the search and track problem based on Random Finite Sets (RFS). Our approach models unknown and known objects using a combined generalized labeled multi-Bernoulli (GLMB) filter. For unseen objects, \ModelNameWithSpace leverages both cardinality and spatial prior distributions, allowing it to operate without prior knowledge of the exact number of objects in the search space. The planner maximizes the mutual information of this unified belief model, creating balanced search and tracking behaviors. We demonstrate our work in a simulated environment, presenting both qualitative results and quantitative improvements 
over a multi-objective method. 
\end{abstract}



\section{Introduction}
\lettrine{T}{here} are numerous applications in which autonomous robots are tasked with searching for or tracking multiple objects. They include scenarios such as disaster relief \cite{Lyu2023, baeck2019drone, AlKaff2019,KARACA2018583}, reconnaissance \cite{Day2021, Moore2021, 8453323}, wildfire management \cite{Akhloufi2021}, and wildlife sciences \cite{Kabir2021, Pfeifer2019,doi:10.1126/scirobotics.abc3000}. In all of these cases, the primary objective is to maximize the knowledge about the objects within the environment, either by increasing the number of discovered objects or by maintaining low uncertainty regarding their current locations.

However, when combining the objectives of discovering undetected objects and tracking detected ones, typical approaches rely on multi-objective formulations that require heuristics or hand-tuned weights to balance competing goals, or they rely on simplifying assumptions. A natural trade-off exists between observing known objects to increase the probability of their existence and reduce state uncertainty, and exploring unknown areas to discover new objects. 
In cases where the exact number of objects is known---or in even simpler scenarios, such as single-object searches like in \cite{Furukawa2012}---the problem can be simplified to a single objective, such as minimizing entropy in a particle filter for each object. However, in real-world applications with multiple objects, knowing the exact count is rare and is more realistically represented by a probability distribution over the number of objects in the environment.

\begin{figure}[!t]
    \centering
\includegraphics[width=.7\linewidth]{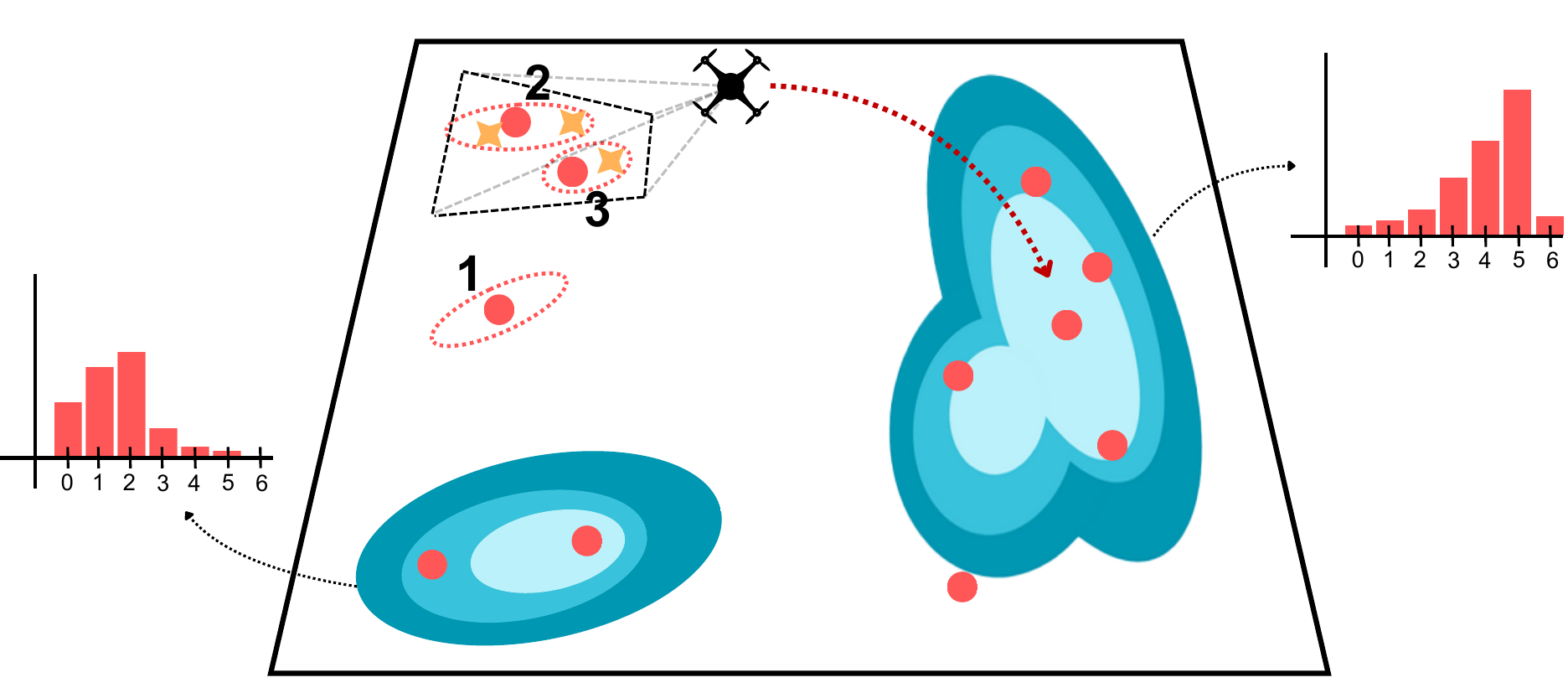}    
\caption{An example scenario of a UAV searching for and tracking objects in a space. The objects are represented by red dots. The UAV is tracking three objects, with two of them currently in the field of view. The Gaussian priors are represented by blue ellipses, and each cluster has an associated cardinality distribution, with the one on the right having a higher cardinality distribution. Using our unified objective function, the UAV planner chooses a path that maximises both tracking and search performance, in this case leaving the tracked targets and searching the area on the right.}
    \label{intro-figure}
\end{figure}

We propose leveraging an object population count distribution in order to create a unified objective function that takes into account the trade-off between gaining observations for known and tracked objects or searching the space for undiscovered objects. 
This is not merely a prior population density distribution across the space, but also a probability distribution of objects we expect within the search space. 
This probability distribution gives more than just an expectation on the total number of objects and is not merely a prior density function or occupancy map over the space. Having a distribution allows for encoding the uncertainty and flexibility based on the confidence of the prior. 
In most search problems, there is a guess on the number and spatial distribution of objects, but the exact count and whereabouts of the objects may not be certain. Examples include finding cattle grazing in a pasture, finding missing hikers, animal censuses, or reconnaissance. 

The contributions of this work are as follows:
\begin{itemize}
    \item[\textbf{\textit{i)}}] A novel belief model formulation for tracked and undiscovered objects through combining them into a single generalized multi-Bernoulli filter (GLMB). 
    \item[\textbf{\textit{ii)}}] An approach that can incorporate both prior cardinality distributions as well as spatial distributions. 
    \item[\textbf{\textit{iii)}}] A unified objective function to search for and track objects using the metric of differential entropy.
    \item[\textbf{\textit{iv)}}] Demonstration of our belief model in a partially observable Markov decision process (POMDP) formulation and presenting planner results with both qualitative and quantitative improvements. 
\end{itemize}

Our method, termed \ModelNameWithSpace (\textit{Unified Search and Track}), is shown to be effective in leveraging the prior beliefs and the cardinality and spatial distributions while still being flexible to inaccurate priors. The combined GLMB filter in our method is effective at handling measurement association, processing the updates for undiscovered and track targets, and outputting object tracks. 
\ModelNameWithSpace does not require choosing parameters to weight search and track objective functions and create intuitive paths that utilize the given priors.

This paper is organized as follows: Section \ref{sec:related_work} gives an overview of related work for searching and tracking objects. Section \ref{sec:problem} defines our problem statement. We present the relevant background for random finite set (RFS) filtering and our POMDP formulation in Section \ref{sec:background}. Section \ref{sec:approach} details \ModelNameWithSpace and implementation. The results are shown and analyzed in Section \ref{sec:results}, and Section \ref{sec:conclusion} is our conclusion and future work.

\section{Related Works}
\label{sec:related_work}

The problem of Multi-Object Tracking (MOT) seeks to provide tracks for an unknown number \hoa{of} objects. When measurements are labeled and data association can be ignored, object states are commonly estimated using the Kalman filter \cite{zhou2023ratt}. Otherwise, as in most multi-object tracking applications, the Kalman filter cannot be directly applied due to the necessity of computing data association. Alternative filtering approaches that handle the data association problem have been proposed, such as the Joint Probability Data Association (JPDA) \cite{fortmann1980jpda}, Multiple Hypothesis Tracker (MHT) \cite{blackman2004mht}, and, more recently, RFS-based frameworks, such as the Probability Hypothesis Density (PHD) filter \cite{mahler2003phd} and the Generalized Labeled Multi-Bernoulli (GLMB) filter \cite{vo2013glmb}.


Random finite sets frameworks model both the number of objects and the object's states as random variables. This makes them suitable for applications where the number of objects is unknown. Previous works have used the RFS-based filters to guide the search to the point of maximum likelihood of finding objects in the local neighborhood \cite{sung2021gmphd-limited-fov} or use greedy search strategies \cite{dames2017detecting, hoa2023multiobjective}, but these methods lack the ability to simultaneously optimize search and track. 

\begin{figure*}[th]
    \centering
\includegraphics[width=\linewidth]{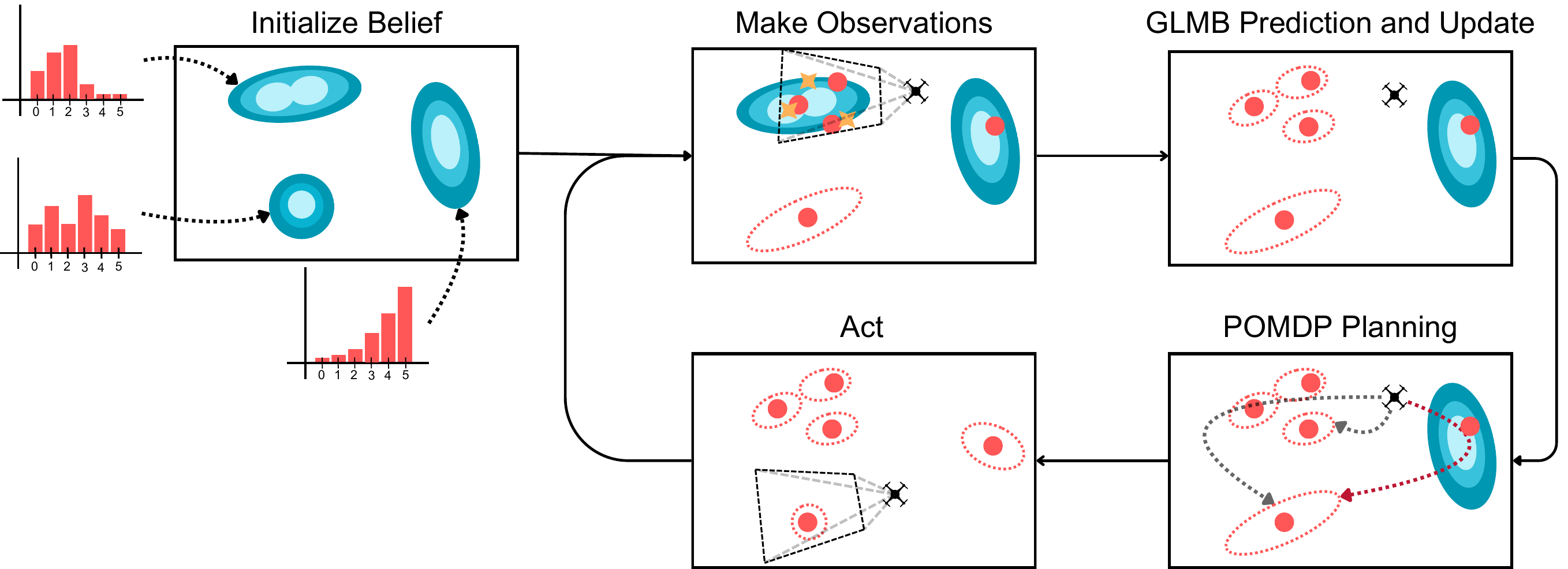}    
\caption{An overview of the \ModelNameWithSpace pipeline. The first step is the belief initialization from the population count, shown on the left. The algorithm then enters the execution loop: observe, update belief, plan, and execute plan. The measurements are represented by purple crosses and the tracks are represented by red dots. With our unified approach, the highest reward plan, highlighted in red, visits the blue density to try to find new objects before visiting one of the tracks to update its state estimation.}
    \label{fig:approach_pipeline}
\end{figure*}


When planning agent paths to maximize the results for MOT, many of the approaches handle simplified cases such as searching for one target \cite{coffin2022}, assuming the number of objects is known and starting with initial tracks for each object in the space \cite{jeong2021}, or searching until an object is found and then switching to purely tracking \cite{Papaioannou2021}. This leads to a simpler problem where there does not have to be a tradeoff between exploring unobserved areas or exploiting the current object tracks. However, these methods would not work well when these assumptions are not met and the number of targets is unknown. 


Other methods work with a fully unknown number of targets. The work by \cite{dames2017detecting} initializes the RFS filter with a prior belief over the space to guide an initial search, which then usually switches to tracking behavior once a target is found. The work of \cite{YOUSUF202293} uses an SMC-PHD filter to estimate object locations and has a tunable parameter that weighs between refining object tracks or searching the space, where search rewards changes based on time and how well the space was seen. There are similar approaches that use occupancy grids or other models for the search space and then try to maximize the objective function for search as well as the one for track \cite{hoa2023multiobjective}. There are others that use combinations of objective functions weighted by scalar values \cite{yan2021}. One way to handle multiple objectives is to use the Pareto set and the global criterion method such as in \cite{hoa2023multiobjective}. However, these methods still necessitate formulating two sometimes disparate objective functions that may not necessarily directly correlate to better performance on the final metrics. 




The question that remains unclear is how one can solve the more general search and track problem, where the exact number of objects may be unknown, and generalize this solution to different scenarios. Although multi-objective search and track has been shown to be able to handle this more general case, it requires hand-picking of the scalar weights~\cite{yan2021} or utilising two conflicting objectives~\cite{hoa2023multiobjective}. 

In this work, we propose a \textit{unified objective} approach that unifies the two objectives (i.e., searching and tracking) and can be applied to different scenarios without modifications. \ModelNameWithSpace leverages prior information, often available to the user in real-world applications, to construct a single belief model that contains objects that were found and objects that are expected to be found based on spatial and cardinality distributions. With the searching and tracking being unified, object measurements can be directly associated with the prior rather than updating the search and track beliefs in a decoupled manner. We then plan on this unified belief model, calculating the value of an action for both search and track and simultaneously optimizing for the two objectives.

%


\section{Problem Statement}
\label{sec:problem}

We address the problem of a single robot tasked with searching and tracking multiple objects (also referred to as targets) with a limited field of view of the workspace. The robot is able to localize itself in the environment, \textit{e.g.} using GPS. The number of objects in the workspace is not known exactly beforehand, but we assume we have access to a population count probability distribution and spatial distribution that we can use to draw information on the expected number of objects and their probable locations. Our goal is to find all objects in the workspace while minimizing the uncertainty of their states. An overview of \ModelNameWithSpace is seen in Fig.~\ref{fig:approach_pipeline}.

\textbf{Robot:} The robot (also referred to as agent) follows a discrete time motion model

\begin{equation*}
u_{k+1} = \tau(u_k, a_k),
\end{equation*}
where $u$ is the robot state and $a \in \mathcal{A}$ is the control input (i.e., robot/agent action). $\mathcal{A}$ is a discrete set of control inputs. 

\textbf{Tracks:} Let $X$ be the multi-object state, 
unknown to the robot. Each object $x \in X$ follows a discrete-time stochastic motion model

\begin{equation*}
    x_{k+1} = f(x_k) + v_k,
\end{equation*}
where $v$ is zero-mean additive Gaussian noise $v \sim \mathcal{N}(\cdot; 0, P)$.

\textbf{Measurements:} We assume that we receive a set $Z_k = \{z_{1,k},\dots,z_{n,k} \}$ 
of position measurements at each time step. The measurement set $Z_k$ is a disjoint union of clutters (i.e., false positives) and detected objects. Each object $x \in X$ has a probability of $P_D(x,u)$ of being detected by the robot with state $u$ and generates a measurement $z \in Z$, given by 
\begin{equation*}
    z_{k} = q(x_k, u_k) + w_k,
\end{equation*}
with additive zero-mean Gaussian noise $w \sim \mathcal{N}(\cdot; 0, \Sigma)$. 

\textbf{Remark 1:} The measurement set $Z$ always depends on the robot state $u$ and the multi-object state $X$. For notational simplicity, we denote $P_D(x) \equiv P_D(x,u)$, $q(x) \equiv q(x,u)$.    


\section{Background}
\label{sec:background}

\subsection{POMDP for Searching and Tracking}

The RFS framework is an effective method to incorporate the stochasticity in target measurements and the multiple track hypotheses as a result. We use a POMDP formulation to plan over this partially-observable problem and define our POMDP tuple as
\begin{itemize}
    \item $\mathbb{X}$ is the space of  multi-object states,
    \item $\mathcal{A}$ is the set of all agent actions,
    \item $\mathbf{\mathit{f}}$$(\mathbf{X}|\mathbf{X}', a')$ is the set of conditional transition probabilities of the agent for $\mathbf{X}, \mathbf{X}' \in \mathbb{X}$ and action $a' \in \mathcal{A}$ is the previous action,
    \item $\mathbb{Z}$ is the observation space,
    \item $\mathbf{\mathit{g}}$$(Z|\mathbf{X}, a')$ is the conditional observation probability, or likelihood, of the observation $Z \in \mathbb{Z}$ from $\mathbf{X} \in \mathbb{X}$ after the action $a' \in \mathcal{A}$,
    \item $\mathcal{R}(Z,\mathbf{X}, a')$ is the reward for the observation $Z \in \mathbb{Z}$ from $\mathbf{X} \in \mathbb{X}$ after the action $a' \in \mathcal{A}$, and
    \item $H$ is the time horizon for the look-ahead.
\end{itemize}


The system evolves according to the transition density $f(\mathbf{X}|\mathbf{X'}, a')$, that coordinates how the multi-object state $\mathbf{X} \in \mathbb{X}$ is updated given the previous state $\mathbf{X'} \in \mathbb{X}$ and the current agent's action $a' \in \mathcal{A}$. After taking action $a \in \mathcal{A}$, the agent obtains an observation $Z \in \mathbb{Z}$ that has likelihood $g(Z|\mathbf{X},a)$. The agent does not have access to the true multi-object state, instead, it uses these observations to update a belief of the true state.

The goal is to find a sequence of actions $(a_k, \dots, a_{k+H-1})$ that will lead to the highest cumulative reward over the time horizon. For that, a value function $V(a_k:a_{k+H-1})$ is constructed from the belief state and cumulative reward over $H$ and assessed to pick the best action sequence. In general, however, this value function does not have an analytical form and numerical approximations are shown to be intractable \cite{hauskrecht2000value}. A tractable approximation can be computed based on the idea of a Predicted Ideal Measurement Set (PIMS) \cite{mahler2007statistical}. Using the multi-object density as the information state $\bm \pi_{a_k-1, k}(\cdot | Z_k)$ at time $k$, we first compute the predicted multi-object density in the next time step
\begin{equation*}
    \bm \hat \pi_{a,j}(\mathbf{\mathbf{X}_j}) = \int \prod_{i=k+1}^j f_i(\mathbf{X}_i|\mathbf{X}_{i-1}, a) \bm \pi_{a_{k-1}, k}(\mathbf{X}_k) \delta \mathbf{X}_{k:j-1} 
\end{equation*}
from which the multi-object state $\hat{\mathbf{X}}_{a,j}$ can be computed via an estimator. To make the computation of the value function tractable, we then assume that \textit{i)} we will apply the same action $a$ over the horizon $k:k+j-1$, and \textit{ii)} we will obtain an ideal set of measurements $Z_{a,j}$ (without noise, clutters) at each time step after applying the action $a$. The approximated value function $V(a)$ is then computed by 
\begin{equation*}
    V(a) =  \sum_{j=k+1}^{k+H} \mathcal{R}(Z_{a,j}, \mathbf{X}_j, a)
\end{equation*}
where $Z_{a,j}$ is the ideal measurement calculated using the measurement model and the current belief of multi-object states.




\subsection{Labelled Multi Bernoulli RFS}

The RFS framework models multi-object states and measurements as RFS-valued random variables. The realizations of such random variables are sets in which both the number of elements (in our case, the number of objects) and the elements themselves (for us, the objects' states) are random. In this work, we adopt the labeled extension of RFS which assigns a unique label to each object.

A Bernoulli RFS $\mathit{X}$ on $\mathbb{X}$ follows a Bernoulli distribution: either has one element with probability $r$, or is empty with probability $1 - r$. Its distribution is given by

\begin{equation}
\label{lmb-rfs}
\pi(\mathit{X}) = \begin{cases}
                    1 - r, & \mathit{X} = \emptyset \\
                    r \, p(x), & \mathit{X} = \{x\}.
                  \end{cases} 
\end{equation}

In the case of multi-object tracking, it means that an object either exists with probability $r$, in which case its state distribution is $p(x)$, or it does not exist with probability $1 - r$. A Labeled Bernoulli RFS is a Bernoulli RFS in which the state $x \in \mathbb{X}$, where $\mathbb{X}$ is the kinematic space, is extended by a label $l \in \mathbb{L}$ to construct a \textit{labeled} state $\mathbf{x} = (x, l)$. A Labeled Multi-Bernoulli RFS (LMB RFS) is the union of multiple independent Labeled Bernoulli RFSs.



\subsection{Multi-Object Dynamic Model}
The evolution of the RFS of objects being tracked is described by the transition density $f_+$. In each time step, an object with state $\mathbf{x}=(x,l)$ either survives to the next time step with probability $P_S(x)$ or dies with probability $1 - P_S(x)$. In case of survival, the object's new state $\mathbf{x}_+$ is given by the transition density $f_s(x_+|x)\delta_l[l_+]$, where $\delta_l[l_+]$ makes sure the label stays the same. New objects can also be born with probability $P_B(l_+)$. In this case, the object's state is $\mathbf{x}_+ = (x_+, l_+)$ and its unlabeled state $x_+$ is distributed according to $f_B(\cdot, l_+)$. It is often assumed that objects survive/die or are born in the next step independently. The transitions of objects to the next time step are independent of the agent's actions.

\subsection{Multi-Object Observation Model}
The measurements and associated noise (misdetections, false alarms and data association uncertainty) are captured by the multi-object observation likelihood $g$. An object with state $\mathbf{x} = (x, l)$ that is in the sensor's field of view is either detected with probability $P_D(x)$, in which case it generates an observation $z$ with likelihood $g(z|x)$, or missed with probability $1 - P_D(x)$. An observation $z$ can also be generated by false alarms, modelled as a Poisson distribution with rate $\kappa(z)$. It is often assumed that detections are independent of each other and clutter. The observations obtained by the agent, including ones generated by objects and false alarms, make the observation set $Z$.

In each time step of the filtering process, all the combinations of measurement-to-object associations are considered. This is needed because data association is unknown. An association map $\gamma$ is a tuple of 1-1 maps

\begin{equation*}
    \gamma: \mathbb{L} \rightarrow \{-1:|Z|\},
\end{equation*}
where $\gamma(l) = - 1$ if object $l$ is not in the RFS (dead or unborn), $\gamma(l) = 0$ if the object was misdetected, and $\gamma(l) > 0$ if the object was detected. It is assumed that each object generates a single measurement, which is guaranteed by the 1-1 property. 

\subsection{GLMB Filter}
The GLMB filter is a rigorous Bayesian filter for multi-object tracking that, under the multi-object system model described above, recursively updates a density of tracked objects in the form of a GLMB.
The GLMB density used in this paper has the form: 
\begin{equation}
    \label{glmb-density}
    \bm \pi(\mathbf{X}) = \sum_{I, \xi} w^{(I, \xi)}\delta_I(\mathcal{L}(\mathbf{X})) \left[p^{(\xi)}\right]^{\mathbf{X}}
\end{equation}
where $p^{\mathbf{X}} = \prod_{\mathbf{x} \in \mathbf{X}} p(\mathbf{x})$ is the joint multi-object distribution, and $p(\cdot, l)$ is a distribution in $\mathbb{X}$; $\mathcal{L}(\mathbf{X})$ gives the labels in $\mathbf{X}$; each $I$ is a finite subset of the label space $\mathbb{L}$; each $\xi$ is a history $\gamma_{1:k}$ of association maps up to the current time; and each $w^{(I, \xi)}$ is a non-negative weight such that $\sum_{I, \xi} w^{(I, \xi)}=1$. For convenience, we refer to the GLMB density as the set of components $\bm \pi \triangleq \left \{\left ( w^{(I, \xi)}, p^{(\xi)} \right ) \right \}$.

Before updating the density with new measurements, a prediction step is applied using the Chapman-Kolmogorov equation to account for the evolution of the targets \cite{mahler2007statistical}

\begin{equation}
    \label{glmb-pred}
    \bm \pi(\mathbf{X}_+) = \int f(\mathbf{X}_+|\mathbf{X})\pi(\mathbf{X}_+)\delta(\mathbf{X}).
\end{equation}

The multi-target posterior is then updated with new measurements via Bayes formula \cite{mahler2007statistical}
\begin{equation}
    \label{glmb-updt}
    \bm \pi(\mathbf{X}|Z) = \frac{g(Z|\mathbf{X})\pi(\mathbf{X})}{\int g(Z|\mathbf{X})\pi(\mathbf{X}) \delta \mathbf{X}}
\end{equation}
where the integral in the denominator is a set integral defined as
\begin{equation*}
    \int f(\mathbf{X}) \delta \mathbf{X} = \sum_{n=0}^{\infty} \int_{\mathbb{X}^{n}} f(\{\mathbf{x}_1, \dots, \mathbf{x}_n\})d(\mathbf{x}_1, \dots, \mathbf{x}_n).
\end{equation*}
These two steps are recursively repeated over time. The GLMB is closed under the prediction and the update, therefore, the result of these operations is also a GLMB. In practice, because the number of GLMB components grows quickly, truncation is done by removing components with a very low probability of existence \cite{vo2019multi}.


\section{Approach}
\label{sec:approach}


We present a unified belief model to jointly plan for searching and tracking, leveraging information from a population count distribution. This information is used to initialize an RFS belief of target states and target count. This belief is a joint representation of known and unknown targets that we expect to find. This joint representation is used to compute a differential entropy reward that simultaneously reasons over search and track. 


\subsection{Belief Space Setup}
The belief initialization is done at the beginning of the mission to model unknown targets that we expect to find based on the information provided in the population count distribution. The population count distribution has two parts: one spatial and one discrete. The spatial part $p^X=\prod_{x\in X}p(x)$ is a probability density over the workspace that informs the probability of finding a target at a certain position. This distribution allows the user to guide the search to areas with a higher likelihood of finding targets. If no prior information is available, a uniform distribution can be used. The discrete part $\rho(n)=\text{Pr}(|X|=n)$ is a cardinality distribution (a discrete probability distribution with $n=0,1,2,\dots$ is an integer number) 
that informs us of the probability of the number of targets in the workspace.

On initialization, we set up the belief space leveraging the information in the population count distribution. 
We first extract the highest possible cardinality in the population count distribution, which we call $N$. In other words, $N$ is the only integer that satisfies $\text{Pr}(|X|>N) = 0$. 
It then allows us to initialize the GLMB density to match the population count distribution. The first step is to initialize the hypotheses. Each hypothesis is a set containing the labels of the tracks in this hypothesis. Let $i_k \in \mathcal{I}$ be the set of all hypotheses with $k$ tracks and $\mathcal{I}$ is the set of all GLMB hypotheses. Formally,
\begin{equation*}
    i_k = C(k, N),
\end{equation*}
where $C(a, b)$ is a function that outputs the set of all $a$ choose $b$ combinations. Considering all combinations is important because the population count does not inform us on which tracks exist---it gives us a probability distribution over the number of tracks, but no information about the exact labels $l_i$. For instance, if we have a total of $4$ tracks in the population count distribution, possible hypotheses with $3$ tracks are $\{l_1, l_2, l_3\}, \text{ or } \{l_1, l_2, l_4\}$, and so on. At the end, the set of all hypothesis $\mathcal{I}$ is the union of the sets $i_k,~\forall~k=1,\dots,N$ tracks,
\begin{equation*}
    \mathcal{I} = i_0 \, \bigcup \, i_1 \, \bigcup \, \dots \, i_j \, \dots \, \bigcup \, i_N.
\end{equation*}
The weight $w(i_k)$ of a hypothesis $i_k$ with $k$ elements is equal to the probability of $k$ elements existing, 
\begin{equation*}
    w(i_k) = \rho(k) = \text{Pr}(|X|=k).
\end{equation*}

Once we initialize all hypotheses, we must initialize the states of the tracks in the hypothesis. Since all hypotheses are combinations of up to $N$ elements, we must initialize $N$ tracks. This is done using the spatial portion of the population count distribution $X$. For each, track $l_i, i = 1, \dots, N$, we sample $N_p$ realizations of $X$ to create a particle cloud that represents the distribution over this track's state $p^{(l_i)}(x)$. We use a Sequential Monte Carlo (SMC), or particle cloud, representation of the track state distribution to allow for multi-modal priors.

At the end of this stage, the GLMB density is a representation of the unknown targets that we expect to find during the search. Multi-object tracking is then performed over this same density, which means that we have a single RFS containing the information used to optimize both search and track.

\subsection{Multi-Object State Estimation}
Multi-object state estimation is done using a combination of the Gaussian Mixture GLMB (GM-GLMB) and the Sequential Monte Carlo GLMB (SMC-GLMB) filters. They only differ in that the joint multi-object distribution in Eq. \ref{glmb-density} is represented by a mixture of Gaussians in the GM-GLMB and by a particle cloud in the SMC-GLMB. The GM-GLMB is used to track objects that have already been found, \textit{i.e.}, it is used for tracking. The SMC-GLMB is used to maintain the belief of objects expected to be found, as described above, \textit{i.e.}, it is used for search. We use a combination of both methods because, while the GM-GLMB is much more efficient, the SMC-GLMB allow us to handle multi-modal priors for search.

After we initialize the belief with the SMC-GLMB, the GLMB density is updated as usual using the recursion described in Eqs. \ref{glmb-pred}-\ref{glmb-updt}, with the addition of negative observations. Negative observations are used to update the belief when we \textit{don't} receive a measurement for an object we predicted could be there: if we don't see an object there, we might not know where it is, but we know we either misdetected it or it is not there. The observation likelihood is thus given by

\begin{equation}
    \label{glmb-updt-ngt}
    g(z_j|\mathbf{x}) = 
    \begin{cases}
        \dfrac{P_D(x)\mathcal{N}(z_j;q(x), \Sigma)}{\kappa(z_j)} &\text{, if } \gamma(\mathcal{L}(\mathbf{x})) = j > 0 \\
        1 - P_D(x) &\text{, if } \gamma(\mathcal{L}(\mathbf{x})) = 0
    \end{cases}.
\end{equation}

When an SMC track's uncertainty is below a threshold set by the user, meaning that an object expected to be found was indeed found, it is converted into a GM track. Potential mismatches (underestimates/overestimates) between the prior population count distribution and the actual number of objects in the workspace are automatically handled by the filter. In the case of underestimates, objects that were not expected to be found are added to the belief by the birth model. Because our agent has a limited field of view of the workspace, our birth model is distributed only across the agent's current field of view. In the case of overestimates, the filter's truncation operation removes SMC tracks when their  probabilities of existence gets too low, meaning they were expected but never found.


\subsection{Planning}
This section gives an overview of our planning approach to jointly search for and track objects in an environment. We explain our use of differential entropy for both the SMC-GLMB and GM-GLMB, and outline our POMDP planning framework.


\subsubsection{Differential Entropy} 


Differential entropy provides a way of measuring the uncertainty of a continuous random variable. In multi-object tracking, it can be used as a measure of the uncertainty of multi-object states.
While differential entropy is intractable to labelled RFS in general, it has an analytical solution to the LMB RFS, given by the set of components $\bm \pi \triangleq \left \{ \left ( r^{(l)}, p^{(l)} \right ) \right \}_{l \in \mathbb{L}}$, where $r^{(l)}$ is the existence probability of track $l$, and $p^{(l)}$ is the state of track $l$. As a result, we propose approximating the GLMB density by an LMB density, which preserves the first moment of the original GLMB density. 
As defined in \cite{hoa2023multiobjective}, the differential entropy for an LMB $\mathbf{X}$ is 
\begin{equation}
\label{eq:diff-ent}
    h(\mathbf{X}) = - \sum_{l\in \mathbb{L}}\left[ r^{(l)}\ln{r^{(l)}} + \tilde{r}^{(l)}\ln{\tilde{r}^{(l)}} + r^{(l)} \langle p^{(l)}, \ln(Kp^{(l)}) \rangle \right]
\end{equation}
where  $\tilde{r}^{(l)} = 1 - r^{(l)}$, $\langle a,b \rangle = \int a(x)b(x) dx$ is the inner product, and $K$ is the hyper-volume unit on $\mathbb{X}$. For the GM-GLMB, this inner product is defined as $\langle p^{(l)}, \ln(Kp^{(l)}) \rangle = 0.5\log[\det(2 \pi \textrm{e} \Sigma^{(l)})]$, where $\Sigma^{(l)}$ is the covariance of $p^{(l)}$, and for the SMC-GLMB it is given by the dot product $\langle p^{(l)}, \ln(Kp^{(l)}) \rangle = [\omega^{(l)}]^{\dagger} \cdot \log(r^{(l)} \omega^{(l)})$, where $\omega^{(l)}$ is the column vector of particle's weights of track $l$ and $\dagger$ denotes a vector/matrix transpose operation. Using this definition of differential entropy, we can define the mutual information between the multi-object RFS $\mathbf{X}$ and measurement RFS $Z$ as

\begin{equation}
    I(\mathbf{X}; Z) = h(\mathbf{X}) - h(\mathbf{X}|Z).
\end{equation}

Mutual information gives a measure of how much information is gained about a random variable by observing another. Therefore, in the context of state estimation where our goal is to choose actions that result in measurements that maximize the mutual information $I(\mathbf{X}; Z)$, which is equivalent to minimising the differential entropy $h(\mathbf{X}|Z)$ since $h(\mathbf{X})$ does not depend on the control action $a$. 

\subsubsection{Planner}

Planning is done in a typical POMDP planning framework, where the goal is to maximize the cumulative reward over a finite horizon $H$. The reward $\mathcal{R}(Z, \mathbf{X}, a)$ is the mutual information between the current belief and observed measurements. We define our value function as

\begin{equation}
    V(a) = \sum_{j=k+1}^{k+H} I(\mathbf{X}_j; Z_j)
\end{equation}
so as to maximize the mutual information between the current belief $\mathbf{X}$ and $Z$, where $Z$ is the PIMS measurement set, which is equivalent to maximizing:
\begin{equation}
    V(a) = -\sum_{j=k+1}^{k+H} h(\mathbf{X}_j| Z_j). 
\end{equation}

While we use a GLMB density as the multi-object state density, the differential entropy in Eq. \ref{eq:diff-ent} is defined for an LMB density. The first step in the planning step is therefore to convert a GLMB into an LMB \cite{reuter2014lmb}. Since our belief $\mathbf{X}$ already contains the expected targets and their states from initialization, and new tracks are always added to this same belief, we only need reason over one value function. When minimizing the differential entropy  $h(\mathbf{X}|Z)$, the planner factors in the differential entropy of the SMC-GLMB in the sense of discovering a new target or reducing uncertainty through negative observations, and the differential entropy of the GM-GLMB tracks that have already been found. The plan thus minimizes the joint uncertainty of search and track.

\section{Results}
\label{sec:results}




To demonstrate the efficacy of \ModelName, we compared its performance against the multi-objective approach in \cite{hoa2023multiobjective}. This baseline uses the global criterion method to handle the objective of minimizing differential entropy for the tracks and maximizing the probability of discovering a target in the occupancy grid. 

We tested both methods in simulation in a handful of different scenarios with different prior distributions to demonstrate that \ModelNameWithSpace is able to consistently leverage the prior information despite its form. All tests were performed with a single agent operating in an environment of $1216$ m $\times$ $1230$ m for a total time of $T = 500$ s. We provide comparisons of the average optimal sub-pattern assignment (OSPA$^2$) distance \cite{beard2017ospa2} achieved by \ModelNameWithSpace and the baseline at the end of the mission across $100$ Monte Carlo (MC) runs.

\subsection{Experiments Settings}

\begin{figure}[!t]
    \centering
    \begin{subfigure}[b]{0.02\textwidth}
        \centering
        \includegraphics[trim={4.5cm 8.3cm 16.3cm 8.5cm},clip,width=\textwidth]{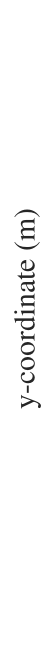}
    \end{subfigure}
    \begin{subfigure}[b]{0.30\textwidth}
        \centering
        \includegraphics[trim={5.3cm 8.3cm 5.3cm 8.95cm},clip,width=\textwidth]{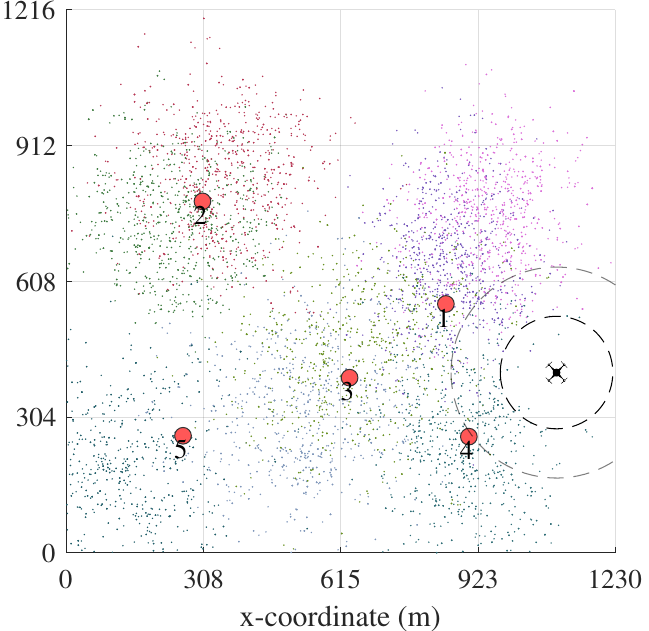}
    \end{subfigure}
    \hspace{\fill}
    \begin{subfigure}[b]{0.30\textwidth}
        \centering
        \includegraphics[trim={5.3cm 8.3cm 5.3cm 8.95cm},clip,width=\textwidth]{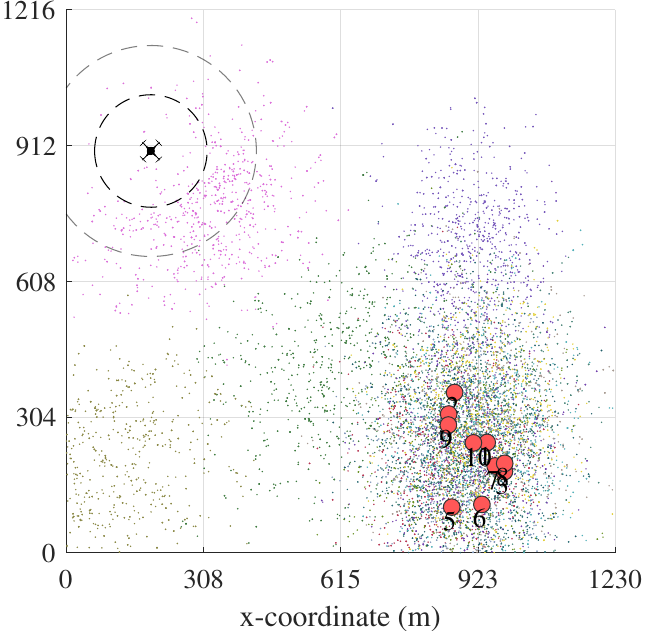}
    \end{subfigure}
    \hspace{\fill}
    \begin{subfigure}[b]{0.30\textwidth}
        \centering
        \includegraphics[trim={5.3cm 8.3cm 5.3cm 8.95cm},clip,width=\textwidth]{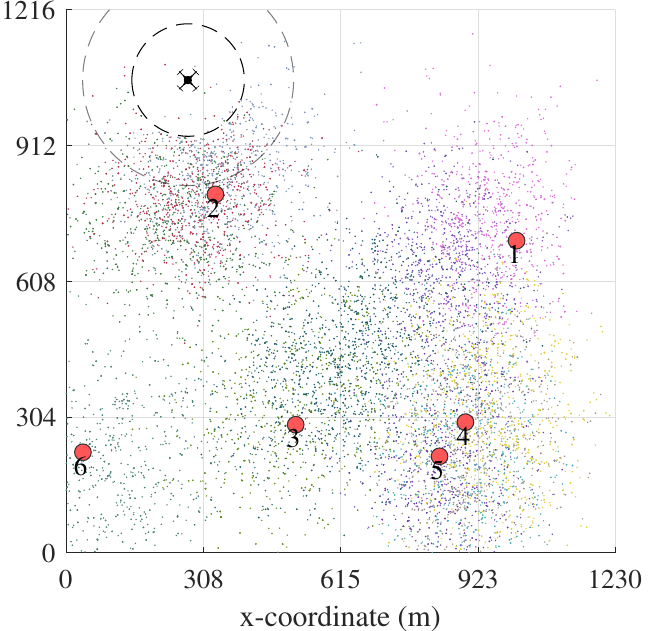}
    \end{subfigure}
    \begin{subfigure}[b]{0.32\textwidth}
        \centering
        \includegraphics[trim={2.2cm 11.3cm 3.9cm 11cm},clip,width=.94\textwidth]{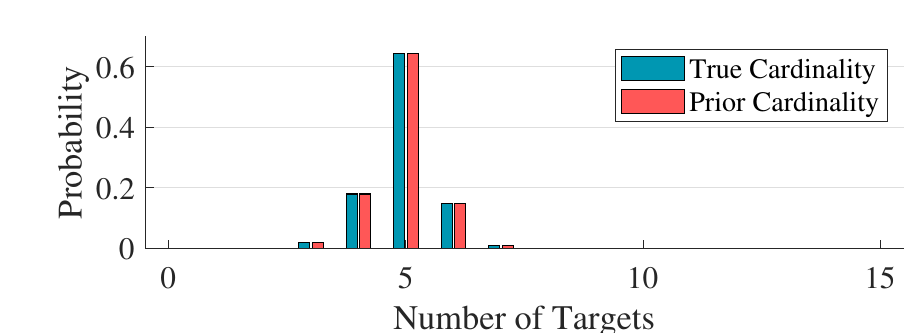}
        \caption{Base Config} 
    \end{subfigure}
    \hspace{\fill}
    \begin{subfigure}[b]{0.32\textwidth}
        \centering
        \includegraphics[trim={2.4cm 11.3cm 3.9cm 11cm},clip,width=.94\textwidth]{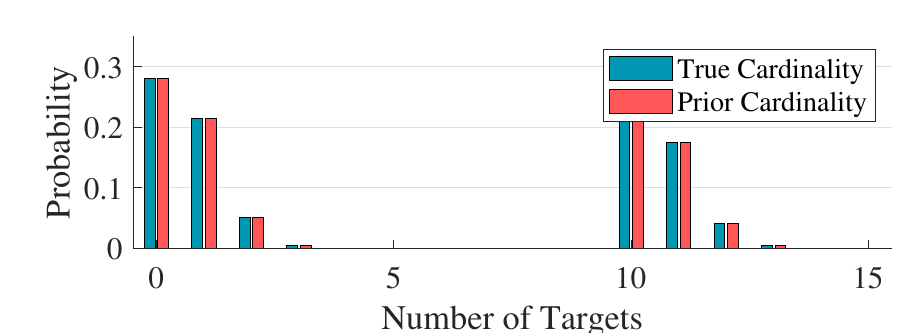}
        \caption{Bimodal} 
    \end{subfigure}
    \hspace{\fill}
    \begin{subfigure}[b]{0.32\textwidth}
        \centering
        \includegraphics[trim={2.6cm 11.3cm 3.9cm 11cm},clip,width=.94\textwidth]{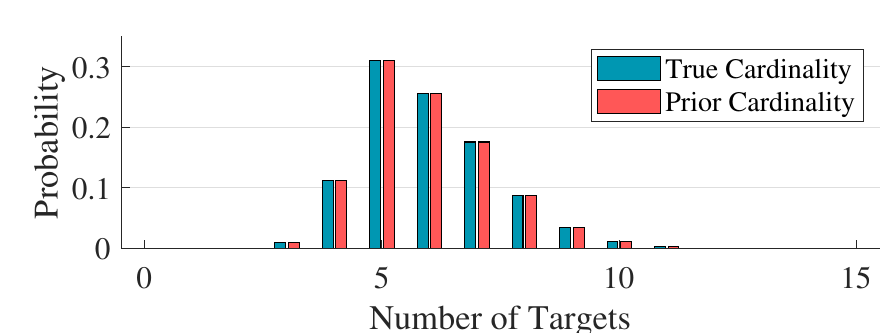}
        \caption{High Variance} 
    \end{subfigure}
    \begin{subfigure}[b]{0.02\textwidth}
        \centering
        \vspace{.5cm}
        \includegraphics[trim={4.5cm 8.3cm 16.3cm 8.5cm},clip,width=\textwidth]{Figures/figure_1.pdf}
    \end{subfigure}
    \begin{subfigure}[b]{0.30\textwidth}
        \centering
        \includegraphics[trim={5.3cm 8.3cm 5.3cm 8.95cm},clip,width=\textwidth]{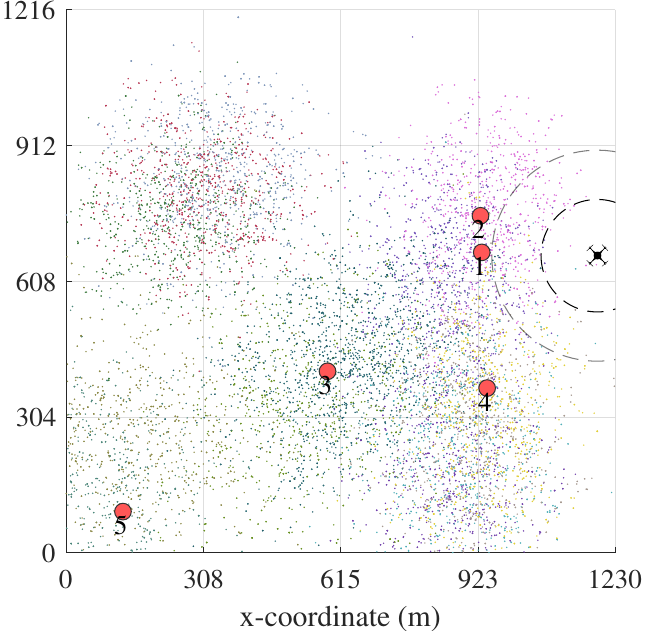}
    \end{subfigure}
    \hfill
    \begin{subfigure}[b]{0.30\textwidth}
        \centering
        \includegraphics[trim={5.3cm 8.3cm 5.3cm 8.95cm},clip,width=\textwidth]{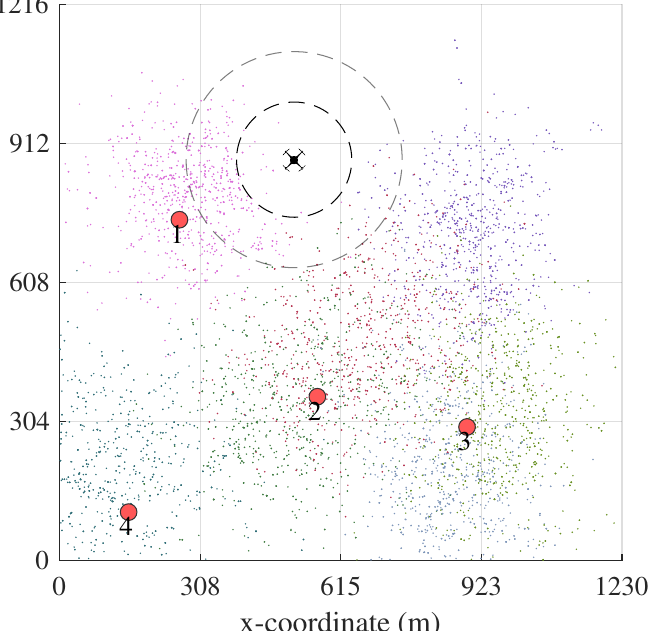}
    \end{subfigure}
    \hfill
    \begin{subfigure}[b]{0.30\textwidth}
        \centering
        \includegraphics[trim={5.3cm 8.3cm 5.3cm 8.95cm},clip,width=\textwidth]{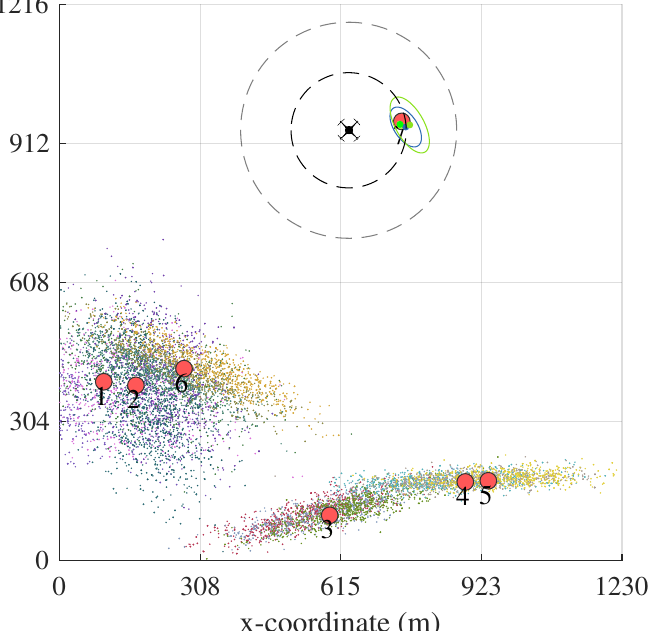}
    \end{subfigure}
    \begin{subfigure}[b]{0.32\textwidth}
        \centering
        \includegraphics[trim={2.2cm 11.3cm 3.9cm 11cm},clip,width=.94\textwidth]{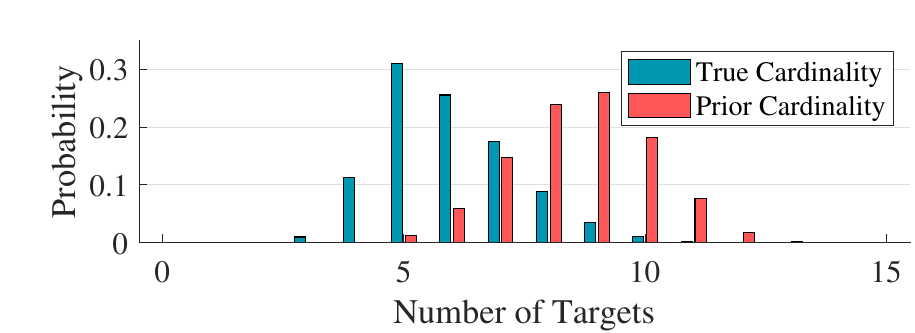}
        \caption{Overestimate} 
    \end{subfigure}
    \hspace{\fill}
    \begin{subfigure}[b]{0.32\textwidth}
        \centering
        \includegraphics[trim={2.4cm 11.3cm 3.9cm 11cm},clip,width=.94\textwidth]{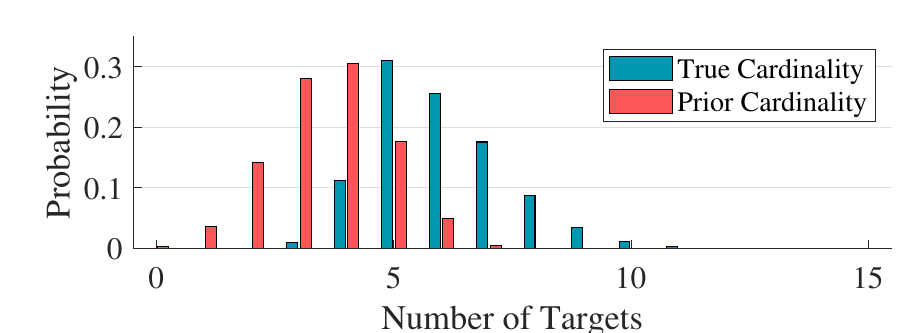}
        \caption{Underestimate} 
    \end{subfigure}
    \hspace{\fill}
    \begin{subfigure}[b]{0.32\textwidth}
        \centering
        \includegraphics[trim={2.6cm 11.3cm 3.9cm 11cm},clip,width=.94\textwidth]{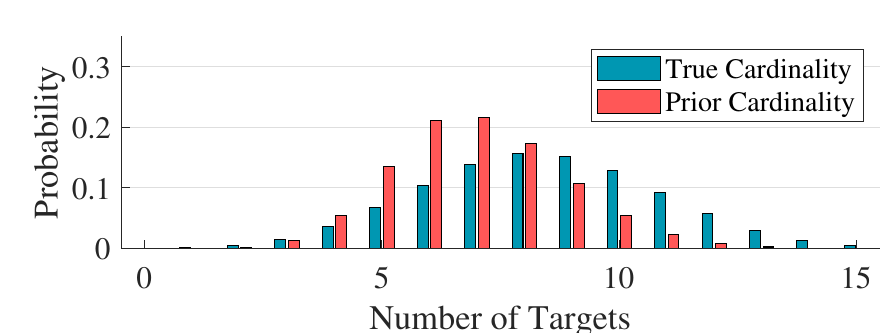}
        \caption{Random} 
        \label{fig:random}
    \end{subfigure}
    \caption{A randomly sampled environment from the five characteristic scenarios and a random environment setup. 
    The true target states are shown in red and the prior SMC particles are plotted in distinct colors.
    Below each environment plot is the true environment cardinality distribution, which is what we sample the ground truth from, and the prior cardinality, which is used for belief initialization.
    }
    \label{fig:prior-setup}
\end{figure}

The initial position of the agent in each MC run was randomly sampled from a uniform distribution over the workspace. Similarly, the initial positions of the targets were randomly sampled from the prior distribution in each run. The targets' states comprise position and velocity $\mathbf{x} = [p_x, p_y, v_x, v_y]^{\dagger}$, 
but for the scope of these tests, we use a random walk motion model for state estimation. The state distributions are represented by particle clouds with a maximum of $1000$ particles per target and $1000$ particles per hypothesis. Our actions space $\mathcal{A}$ comprises $12$ even spaced trajectories emanating outward from the current agent state, with one additional action of staying in the same state. The planning horizon is $40$ steps and each action belief sampling interval is $10$ steps.   

We assume the sensor is subject to noisy observations. We model false alarms as a Poisson distribution with a clutter rate of $\lambda = 0.01$ and account for misdetections assuming that a target within the sensor's field of view is detected with a detection probability of $P_D = 0.8825$. We measure performance using the OSPA$^2$ distance with a cut-off $c = 50$~m. This metric accounts for errors in both the estimated number of objects and their states. Therefore, lower OSPA$^2$ means better performance.



\subsection{Results and Discussion}

We first create a set of characteristic scenarios to demonstrate various environmental conditions and the effect of the prior on the behavior and performance of \ModelName. \textit{Base Config}: Our base configuration is to compare against the subsequent variations. There are five normal distributions of targets placed throughout the environment. In the base configuration, the prior distributions match the true distributions. \textit{Bimodal}: In this scenario, the spatial distributions stay the same, but the cardinality distributions are altered to create a population cardinality distribution that is bimodal. \textit{High Variance}: This scenario has a higher spread in the cardinality distributions than the base configuration. \textit{Overestimate}: We keep the same cardinality distribution as the high variance case, but we test \ModelName's performance when the prior cardinality distribution is an overestimate. The true expected target count is 5.95, but the prior has an expected target count of 8.63. \textit{Underestimate}: This case tests when the prior is an underestimate. The true expected target count again is 5.95, but the prior has an expected target count of 3.60.


A random environment sampled for each of these scenarios is shown in Fig.~\ref{fig:prior-setup}. We also visualize the cardinality distributions for each of the priors. We note that each distribution has a unique cardinality distribution rather than one combined distribution for the total population. Each test randomly samples the true target cardinality and location from the true distributions. 

The results for the five scenarios are summarized in Table \ref{tab:method-results}. We can see that \ModelNameWithSpace outperforms the baseline in all of the scenarios. \ModelNameWithSpace performs well in both the base configuration and bimodal case, where there is a strong incentive to first visit clusters with high expected cardinality rather than searching uniformly in the space or the other priors. If a low number of targets are discovered in a prior cluster, the agent may move on to the other priors. Conversely,  if the agent finds a high number of targets, it may choose to revisit the targets many times or even keep them within the field of view to maintain their tracks.

\begin{table}[t]
    \centering
        \caption{Results from 100 MC tests for each characteristic scenario. Table values are the mean OSPA$^2$ error and margin of error for a $95\%$ confidence level.}
    \newlength{\mapspace}
    \setlength{\mapspace}{2pt}
    \begin{tabularx}{0.54\linewidth}{ccc}
       \toprule
       Scenario & Ours~(m) & Multi-Objective \cite{hoa2023multiobjective} (m)\\
       \midrule
        Base Config & $\mathbf{25.01\pm1.14}$  & $32.78\pm0.96$  \\   
       Bimodal & $\mathbf{20.56\pm2.36}$  & $29.04\pm2.35$  \\ 
       High Variance & $\mathbf{28.15\pm1.36}$  & $33.73\pm1.02$  \\ 
       Overestimate& $\mathbf{29.98\pm1.39}$  & $33.93\pm1.02$  \\ 
       Underestimate& $\mathbf{22.35\pm1.03}$  & $33.79\pm1.03$  \\ 
       \bottomrule
    \end{tabularx}
    \label{tab:method-results}
\end{table}

In the cases of high cardinality variance, the performance gap between the methods decreases. This is likely because the baseline search strategy performs well by searching the entire space. With the targets spread out fairly evenly, almost any action would lead to good performance. As expected, when we alter the prior to be an overestimate, the performance gap between the methods decreases. This is due to \ModelNameWithSpace being imbalanced in the searching versus tracking behavior and being biased to search more, as it expects to discover more targets in the space. This shows that \ModelNameWithSpace doesn't fail when the prior is an overestimate, but rather its performance decreases toward being closer to the baseline performance. In the underestimate case, our proposed method performs much better than the baseline. This demonstrates \ModelName's flexibility in discovering unexpected targets without affecting the remaining expected number of targets. Thus, the cardinality distributions do not have to be completely accurate for \ModelNameWithSpace to outperform the baseline and benefit the planner.

\begin{figure}[t]
    \centering
    \begin{subfigure}[b]{0.016\textwidth}
        \centering
        \includegraphics[trim={4.5cm 8.3cm 16.3cm 8.5cm},clip,width=\textwidth]{Figures/figure_1.pdf}
    \end{subfigure}
    \begin{subfigure}[b]{0.24\textwidth}
        \centering
        \includegraphics[trim={5.3cm 8.3cm 5.3cm 8.5cm},clip,width=\textwidth]{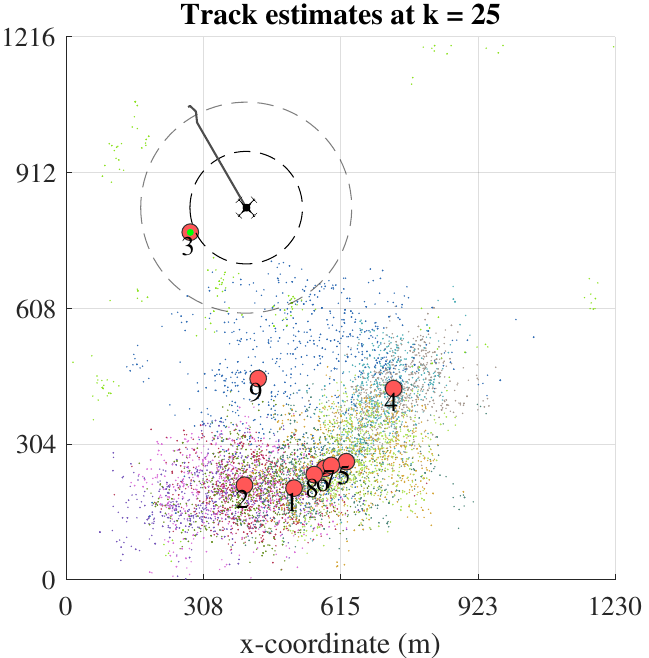}
    \end{subfigure}
    \hfill
    \begin{subfigure}[b]{0.24\textwidth}
        \centering
        \includegraphics[trim={5.3cm 8.3cm 5.3cm 8.5cm},clip,width=\textwidth]{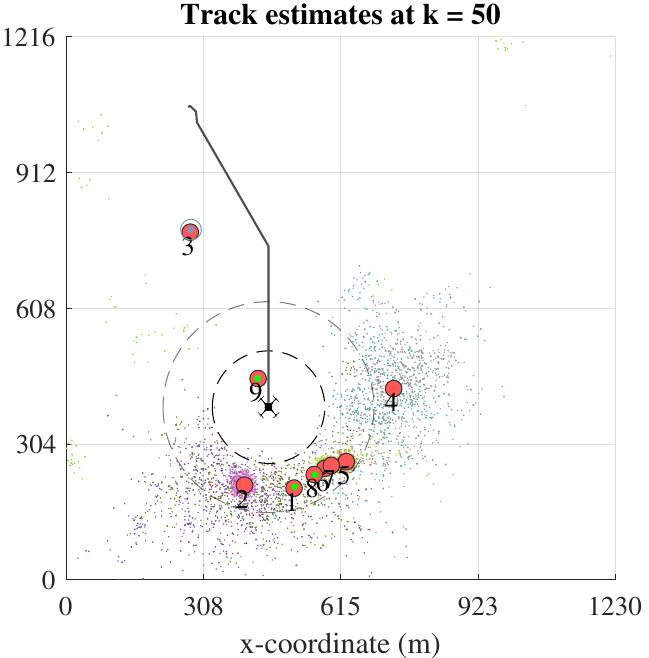}
    \end{subfigure}
    \hfill
    \begin{subfigure}[b]{0.24\textwidth}
        \centering
        \includegraphics[trim={5.3cm 8.3cm 5.3cm 8.5cm},clip,width=\textwidth]{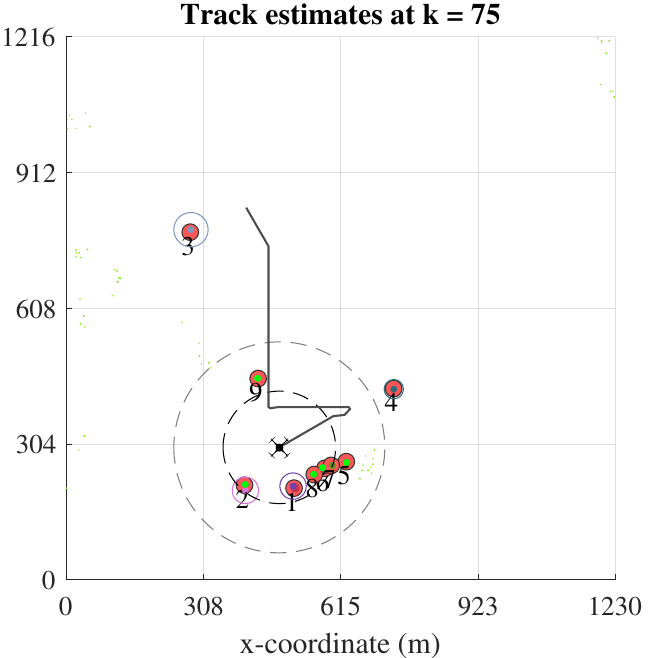}
    \end{subfigure}
    \hfill
    \begin{subfigure}[b]{0.24\textwidth}
        \centering
        \includegraphics[trim={5.3cm 8.3cm 5.3cm 8.5cm},clip,width=\textwidth]{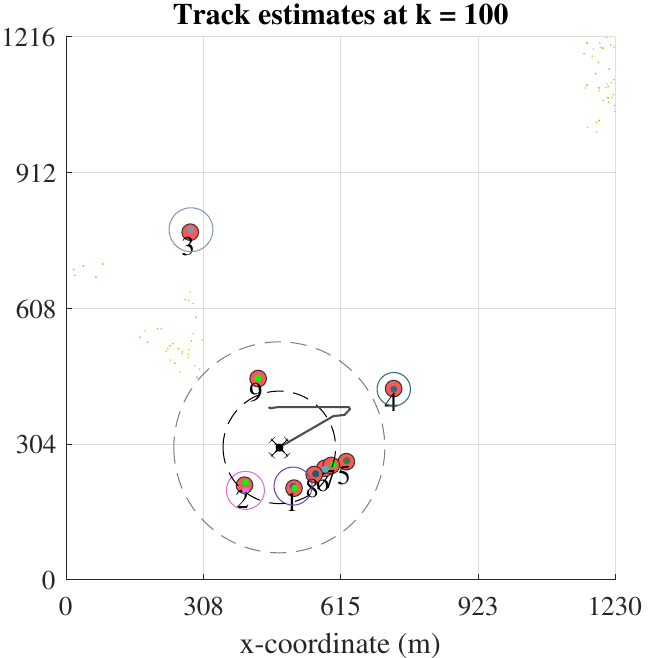}
    \end{subfigure}
    
    \caption{An example run of \ModelNameWithSpace at different time steps in our simulation environment. The range of detection is shown in dotted lines around the UAV, where the probability of detection drops off between black to grey circles. The targets are shown as red dots and the EKF track covariance can be seen around the targets being tracked. SMC belief particles are visualized with each hypothesis having a different color.}
    \label{fig:qualitative2}
\end{figure}

We present an example execution of \ModelNameWithSpace in Fig. \ref{fig:qualitative2}. This example demonstrates the intuitive behavior of the UAV (our considered agent), driven by our unified objective function that leverages the prior spatial and cardinality distributions. The UAV quickly discovers two targets, but rather than staying to track them, it moves on to search the other priors due to the potential to discover a high number of new targets. The UAV effectively discovers all of the targets in the cluster in the bottom left and then chooses to stay in a position where it can receive measurements for a large number of the targets.

We then test \ModelNameWithSpace on an environment with a random number of prior Gaussian clusters, locations, covariance, and cardinality distributions. The number of clusters is between $1-6$, and they are placed randomly within the search space with a $100$ m buffer from the edge. For the covariance of each Gaussian, we sample the standard deviation of the diagonal elements between $0-64$ and for the off-diagonal elements, we sample a correlation coefficient between $-1$ and $1$, which we then multiply by the sampled standard deviations. For the cardinality distributions, we allocate a maximum of $3$ targets per cluster, and we sample probabilities between $0$ and $1$ for each element of the cardinality distribution and then normalize the vector of probabilities. An example of a sampled environment is shown in Fig. \ref{fig:random}.

\begin{figure}[!t]
    \centering
    \includegraphics[trim={0cm 6cm 0cm 6cm},clip,width=.7\linewidth]{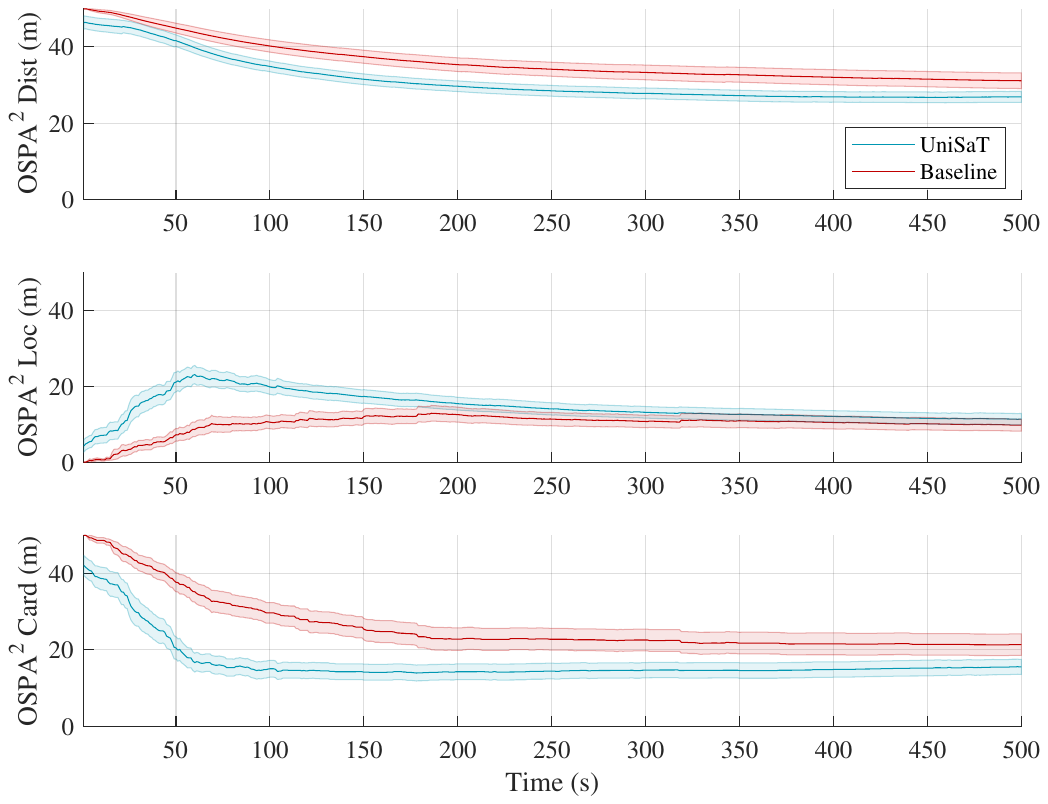}    
    \caption{The mean OSPA$^2$ results of \ModelNameWithSpace and the baseline over 100 MC runs on the environment with random configurations, plotted with the $95\%$ confidence interval.}
    \label{fig:ospa}
\end{figure}

The results for 100 MC tests are shown in Fig. \ref{fig:ospa}. We can see that \ModelNameWithSpace quickly reduces the cardinality error, as it is able to exploit its understanding of the prior and the value of identifying the undiscovered targets left in the search space. As a result of this, the localization error is higher than the baseline in the beginning but decreases to closely match the baseline over the course of the tests. The overall OSPA$^2$ for \ModelNameWithSpace is lower than that of the baseline. The final mean OSPA$^2$ error is $26.89 \pm 1.44$~m for \ModelNameWithSpace and $31.14 \pm 2.05$~m for the baseline, with the $\pm$ values representing the margin of error for a $95\%$ confidence level. 

\section{Conclusion and Future Work}
\label{sec:conclusion}
This work presents \ModelName, a planner and novel unified belief model for searching and tracking with an autonomous agent. The proposed belief model is constructed using prior knowledge of the objects in the environment and allows the planner to simultaneously optimize for searching and tracking, which normally are treated as two competitive objectives in hand-tuned multi-objective approaches. \ModelNameWithSpace can incorporate both the prior spatial and cardinality distributions and even uses them when creating new tracks from measurements. 
We compare \ModelNameWithSpace to a multi-objective approach through Monte Carlo simulations and show we outperform this baseline in a variety of scenarios with different distributions of objects in the environment. 

Future work could choose to strictly adhere to the prior population cardinality distribution and prune all hypotheses that do not fall within a threshold of the distribution. This approach would allow for all measurements to affect all hypotheses in the environment. When an object that was expected to be found in a region would not be found, the prior would be redistributed to other regions to keep the expectations on the number of targets constant. Another avenue could include exploring other methods of encoding the search space that \hoa{is} efficient while still taking into account measurement association and correlations between priors.

\section*{Acknowledgments}
This material is based upon work supported by the National Science Foundation Graduate Research Fellowship under Grant No. DGE1745016. This work is supported by the Office of Naval Research (Grant N00014-21-1-2110), and the Australian Research Council (ARC) Linkage Project No. LP200301507.

\bibliography{references}

\begin{thebibliography}{30}
\newcommand{\enquote}[1]{``#1''}
\providecommand{\natexlab}[1]{#1}
\providecommand{\url}[1]{\texttt{#1}}
\providecommand{\urlprefix}{URL }
\expandafter\ifx\csname urlstyle\endcsname\relax
  \providecommand{\doi}[1]{\discretionary{}{}{}https://doi.org/#1}\else
  \providecommand{\doi}[1]{\discretionary{}{}{}\urlstyle{rm}\url{https://doi.org/#1}}\fi

\bibitem[{Lyu et~al.(2023)Lyu, Zhao, Huang, and Huang}]{Lyu2023}
Lyu, M., Zhao, Y., Huang, C., and Huang, H., \enquote{Unmanned Aerial Vehicles for Search and Rescue: A Survey,} \emph{Remote Sensing}, Vol.~15, No.~13, 2023, p. 3266.
\newblock \doi{10.3390/rs15133266}, \urlprefix\url{http://dx.doi.org/10.3390/rs15133266}.

\bibitem[{Baeck et~al.(2019)Baeck, Lewyckyj, Beusen, Horsten, and Pauly}]{baeck2019drone}
Baeck, P., Lewyckyj, N., Beusen, B., Horsten, W., and Pauly, K., \enquote{Drone based near real-time human detection with geographic localization,} \emph{The International Archives of the Photogrammetry, Remote Sensing and Spatial Information Sciences}, Vol.~42, 2019, pp. 49--53.

\bibitem[{Al-Kaff et~al.(2019)Al-Kaff, Gómez-Silva, Moreno, de~la Escalera, and Armingol}]{AlKaff2019}
Al-Kaff, A., Gómez-Silva, M., Moreno, F., de~la Escalera, A., and Armingol, J., \enquote{An Appearance-Based Tracking Algorithm for Aerial Search and Rescue Purposes,} \emph{Sensors}, Vol.~19, No.~3, 2019, p. 652.
\newblock \doi{10.3390/s19030652}, \urlprefix\url{http://dx.doi.org/10.3390/s19030652}.

\bibitem[{Karaca et~al.(2018)Karaca, Cicek, Tatli, Sahin, Pasli, Beser, and Turedi}]{KARACA2018583}
Karaca, Y., Cicek, M., Tatli, O., Sahin, A., Pasli, S., Beser, M.~F., and Turedi, S., \enquote{The potential use of unmanned aircraft systems (drones) in mountain search and rescue operations,} \emph{The American Journal of Emergency Medicine}, Vol.~36, No.~4, 2018, pp. 583--588.
\newblock \doi{https://doi.org/10.1016/j.ajem.2017.09.025}, \urlprefix\url{https://www.sciencedirect.com/science/article/pii/S0735675717307507}.

\bibitem[{Day and Salmon(2021)}]{Day2021}
Day, R., and Salmon, J., \enquote{A Framework for Multi-UAV Persistent Search and Retrieval with Stochastic Target Appearance in a Continuous Space,} \emph{Journal of Intelligent \& Robotic Systems}, Vol. 103, No.~4, 2021.
\newblock \doi{10.1007/s10846-021-01484-1}, \urlprefix\url{http://dx.doi.org/10.1007/s10846-021-01484-1}.

\bibitem[{Moore et~al.(2021)Moore, Bidstrup, Peterson, and Beard}]{Moore2021}
Moore, J.~J., Bidstrup, C.~C., Peterson, C.~K., and Beard, R.~W., \enquote{Tracking Multiple Vehicles Constrained to a Road Network From a UAV with Sparse Visual Measurements,} \emph{Frontiers in Robotics and AI}, Vol.~8, 2021.
\newblock \doi{10.3389/frobt.2021.744185}, \urlprefix\url{http://dx.doi.org/10.3389/frobt.2021.744185}.

\bibitem[{Moon and Peterson(2018)}]{8453323}
Moon, B.~G., and Peterson, C.~K., \enquote{Learned Search Parameters For Cooperating Vehicles using Gaussian Process Regressions,} \emph{2018 International Conference on Unmanned Aircraft Systems (ICUAS)}, 2018, pp. 493--502.
\newblock \doi{10.1109/ICUAS.2018.8453323}.

\bibitem[{Akhloufi et~al.(2021)Akhloufi, Couturier, and Castro}]{Akhloufi2021}
Akhloufi, M.~A., Couturier, A., and Castro, N.~A., \enquote{Unmanned Aerial Vehicles for Wildland Fires: Sensing, Perception, Cooperation and Assistance,} \emph{Drones}, Vol.~5, No.~1, 2021, p.~15.
\newblock \doi{10.3390/drones5010015}, \urlprefix\url{http://dx.doi.org/10.3390/drones5010015}.

\bibitem[{Kabir and Lee(2021)}]{Kabir2021}
Kabir, R.~H., and Lee, K., \enquote{Wildlife Monitoring Using a Multi-UAV System with Optimal Transport Theory,} \emph{Applied Sciences}, Vol.~11, No.~9, 2021, p. 4070.
\newblock \doi{10.3390/app11094070}, \urlprefix\url{http://dx.doi.org/10.3390/app11094070}.

\bibitem[{Pfeifer et~al.(2019)Pfeifer, Barbosa, Mustafa, Peter, R\"{u}mmler, and Brenning}]{Pfeifer2019}
Pfeifer, C., Barbosa, A., Mustafa, O., Peter, H.-U., R\"{u}mmler, M.-C., and Brenning, A., \enquote{Using Fixed-Wing UAV for Detecting and Mapping the Distribution and Abundance of Penguins on the South Shetlands Islands, Antarctica,} \emph{Drones}, Vol.~3, No.~2, 2019, p.~39.
\newblock \doi{10.3390/drones3020039}, \urlprefix\url{http://dx.doi.org/10.3390/drones3020039}.

\bibitem[{Shah et~al.(2020)Shah, Ballard, Schmidt, and Schwager}]{doi:10.1126/scirobotics.abc3000}
Shah, K., Ballard, G., Schmidt, A., and Schwager, M., \enquote{Multidrone aerial surveys of penguin colonies in Antarctica,} \emph{Science Robotics}, Vol.~5, No.~47, 2020, p. eabc3000.
\newblock \doi{10.1126/scirobotics.abc3000}, \urlprefix\url{https://www.science.org/doi/abs/10.1126/scirobotics.abc3000}.

\bibitem[{Furukawa et~al.(2012)Furukawa, Mak, Durrant-Whyte, and Madhavan}]{Furukawa2012}
Furukawa, T., Mak, L.~C., Durrant-Whyte, H., and Madhavan, R., \enquote{Autonomous Bayesian Search and Tracking, and its Experimental Validation,} \emph{Advanced Robotics}, Vol.~26, No. 5–6, 2012, p. 461–485.
\newblock \doi{10.1163/156855311x617461}, \urlprefix\url{http://dx.doi.org/10.1163/156855311X617461}.

\bibitem[{Zhou and Kumar(2023)}]{zhou2023ratt}
Zhou, L., and Kumar, V., \enquote{Robust Multi-Robot Active Target Tracking Against Sensing and Communication Attacks,} \emph{IEEE Transactions on Robotics}, Vol.~39, No.~3, 2023, pp. 1768--1780.
\newblock \doi{10.1109/TRO.2022.3233341}.

\bibitem[{Fortmann et~al.(1980)Fortmann, Bar-Shalom, and Scheffe}]{fortmann1980jpda}
Fortmann, T.~E., Bar-Shalom, Y., and Scheffe, M., \enquote{Multi-target tracking using joint probabilistic data association,} \emph{1980 19th IEEE Conference on Decision and Control including the Symposium on Adaptive Processes}, IEEE, 1980, pp. 807--812.

\bibitem[{Blackman(2004)}]{blackman2004mht}
Blackman, S.~S., \enquote{Multiple hypothesis tracking for multiple target tracking,} \emph{IEEE Aerospace and Electronic Systems Magazine}, Vol.~19, No.~1, 2004, pp. 5--18.

\bibitem[{Mahler(2003)}]{mahler2003phd}
Mahler, R.~P., \enquote{Multitarget Bayes filtering via first-order multitarget moments,} \emph{IEEE Transactions on Aerospace and Electronic systems}, Vol.~39, No.~4, 2003, pp. 1152--1178.

\bibitem[{Vo and Vo(2013)}]{vo2013glmb}
Vo, B.-T., and Vo, B.-N., \enquote{Labeled Random Finite Sets and Multi-Object Conjugate Priors,} \emph{IEEE Transactions on Signal Processing}, Vol.~61, No.~13, 2013, pp. 3460--3475.
\newblock \doi{10.1109/TSP.2013.2259822}.

\bibitem[{Sung and Tokekar(2021)}]{sung2021gmphd-limited-fov}
Sung, Y., and Tokekar, P., \enquote{Gm-phd filter for searching and tracking an unknown number of targets with a mobile sensor with limited fov,} \emph{IEEE Transactions on Automation Science and Engineering}, Vol.~19, No.~3, 2021, pp. 2122--2134.

\bibitem[{Dames et~al.(2017)Dames, Tokekar, and Kumar}]{dames2017detecting}
Dames, P., Tokekar, P., and Kumar, V., \enquote{Detecting, localizing, and tracking an unknown number of moving targets using a team of mobile robots,} \emph{The International Journal of Robotics Research}, Vol.~36, No. 13-14, 2017, pp. 1540--1553.

\bibitem[{Nguyen et~al.(2023)Nguyen, Vo, Vo, Rezatofighi, and Ranasinghe}]{hoa2023multiobjective}
Nguyen, H.~V., Vo, B.-N., Vo, B.-T., Rezatofighi, H., and Ranasinghe, D.~C., \enquote{Multi-Objective Multi-Agent Planning for Discovering and Tracking Multiple Mobile Objects,} , 2023.
\newblock \doi{10.48550/arXiv.2203.04551}.

\bibitem[{Coffin et~al.(2022)Coffin, Abraham, Sartoretti, Dillstrom, and Choset}]{coffin2022}
Coffin, H., Abraham, I., Sartoretti, G., Dillstrom, T., and Choset, H., \enquote{Multi-Agent Dynamic Ergodic Search with Low-Information Sensors,} \emph{2022 International Conference on Robotics and Automation (ICRA)}, 2022, pp. 11480--11486.
\newblock \doi{10.1109/ICRA46639.2022.9812037}.

\bibitem[{Jeong et~al.(2021)Jeong, Hassani, Morari, Lee, and Pappas}]{jeong2021}
Jeong, H., Hassani, H., Morari, M., Lee, D.~D., and Pappas, G.~J., \enquote{Deep Reinforcement Learning for Active Target Tracking,} \emph{2021 IEEE International Conference on Robotics and Automation (ICRA)}, 2021, pp. 1825--1831.
\newblock \doi{10.1109/ICRA48506.2021.9561258}.

\bibitem[{Papaioannou et~al.(2021)Papaioannou, Kolios, Theocharides, Panayiotou, and Polycarpou}]{Papaioannou2021}
Papaioannou, S., Kolios, P., Theocharides, T., Panayiotou, C.~G., and Polycarpou, M.~M., \enquote{A Cooperative Multiagent Probabilistic Framework for Search and Track Missions,} \emph{IEEE Transactions on Control of Network Systems}, Vol.~8, No.~2, 2021, pp. 847--858.
\newblock \doi{10.1109/TCNS.2020.3038843}.

\bibitem[{Yousuf et~al.(2022)Yousuf, Lendek, and Buşoniu}]{YOUSUF202293}
Yousuf, B., Lendek, Z., and Buşoniu, L., \enquote{Exploration-Based Search for an Unknown Number of Targets using a UAV,} \emph{IFAC-PapersOnLine}, Vol.~55, No.~15, 2022, pp. 93--98.
\newblock \doi{https://doi.org/10.1016/j.ifacol.2022.07.614}, \urlprefix\url{https://www.sciencedirect.com/science/article/pii/S2405896322010254}, 6th IFAC Conference on Intelligent Control and Automation Sciences ICONS 2022.

\bibitem[{Yan et~al.(2021)Yan, Jia, and Bai}]{yan2021}
Yan, P., Jia, T., and Bai, C., \enquote{Searching and Tracking an Unknown Number of Targets: A Learning-Based Method Enhanced with Maps Merging,} \emph{Sensors}, Vol.~21, No.~4, 2021.
\newblock \doi{10.3390/s21041076}, \urlprefix\url{https://www.mdpi.com/1424-8220/21/4/1076}.

\bibitem[{Hauskrecht(2000)}]{hauskrecht2000value}
Hauskrecht, M., \enquote{Value-function approximations for partially observable Markov decision processes,} \emph{Journal of artificial intelligence research}, Vol.~13, 2000, pp. 33--94.

\bibitem[{Mahler(2007)}]{mahler2007statistical}
Mahler, R., \emph{Statistical multisource-multitarget information fusion}, Artech, 2007.

\bibitem[{Vo et~al.(2019)Vo, Vo, and Beard}]{vo2019multi}
Vo, B.-N., Vo, B.-T., and Beard, M., \enquote{Multi-sensor multi-object tracking with the generalized labeled multi-Bernoulli filter,} \emph{IEEE Transactions on Signal Processing}, Vol.~67, No.~23, 2019, pp. 5952--5967.

\bibitem[{Reuter et~al.(2014)Reuter, Vo, Vo, and Dietmayer}]{reuter2014lmb}
Reuter, S., Vo, B.-T., Vo, B.-N., and Dietmayer, K., \enquote{The labeled multi-Bernoulli filter,} \emph{IEEE Transactions on Signal Processing}, Vol.~62, No.~12, 2014, pp. 3246--3260.

\bibitem[{Beard et~al.(2017)Beard, Vo, and Vo}]{beard2017ospa2}
Beard, M., Vo, B.~T., and Vo, B.-N., \enquote{OSPA (2): Using the OSPA metric to evaluate multi-target tracking performance,} \emph{2017 International Conference on Control, Automation and Information Sciences (ICCAIS)}, IEEE, 2017, pp. 86--91.

\end{thebibliography}

\end{document}